\documentclass{article}
\pdfoutput=1

\usepackage{amsmath}
\usepackage{amssymb}
\usepackage{graphicx} 

\usepackage{natbib}

\usepackage{fancybox,graphicx}
\usepackage{float}

\usepackage[accepted]{icml2010} 

\usepackage{stmaryrd,dsfont}

\newenvironment{Ouralgorithm}[1][\  ] %
{
\rm
\begin{tabbing} 
.\=...\=...\=...\=...\=  \+ \kill
} %
{\end{tabbing}
}

\floatstyle{ruled}
\newfloat{Algorithm}{thp}{alg}

\newenvironment{Balgorithm} %
{
\begin{minipage}{1.0\linewidth} \begin{Ouralgorithm} %
}
{ \end{Ouralgorithm} \end{minipage} 
}

\usepackage{rotating}
\usepackage{array}
\usepackage{multirow}
\usepackage{color}

\usepackage{tikz}

\newtheorem{lemma}{Lemma}

\newtheorem{theorem}{Theorem}

\newcommand{\x}{{\mathbf x}}

\newcommand{\z}{{\mathbf z}}
\newcommand{\w}{{\mathbf w}}

\renewcommand{\v}{{\mathbf v}}
\renewcommand{\u}{{\mathbf u}}
\newcommand{\e}{{\mathbf e}}

\newcommand{\sgn}{{\mathrm{sgn}}}

\renewcommand{\r}{{\mathbf{r}}}

\newcommand{\gb}[1]{\boldsymbol{#1}}

\newcommand{\BlackBox}{\rule{1.5ex}{1.5ex}}  
\newenvironment{proof}{\par\noindent{\bf Proof\ }}{\hfill\BlackBox\\[2mm]}

\newcommand{\reals}{\mathbb{R}}

\DeclareMathOperator*{\E}{\mathbb{E}}
\DeclareMathOperator*{\prob}{\mathbb{P}}
\newcommand{\D}{\mathcal{D}}
\newcommand{\inner}[1]{\langle #1 \rangle}
\newcommand{\half}{\frac{1}{2}}
\newcommand{\thalf}{\tfrac{1}{2}}
\newcommand{\eqdef}{\stackrel{\mathrm{def}}{=}}

\DeclareMathOperator*{\argmin}{argmin} 

\renewcommand{\eqref}[1]{(\ref{#1})}
\newcommand{\figref}[1]{Figure~\ref{#1}}
\newcommand{\secref}[1]{Section~\ref{#1}}
\newcommand{\thmref}[1]{Theorem~\ref{#1}}
\newcommand{\lemref}[1]{Lemma~\ref{#1}}

\newcommand\indct[1]{\ensuremath{\mathds{1}\!_{[{#1}]}}}
\newcommand{\Ej}{\mathds{E}_j}
\newcommand{\Eu}{\mathds{E}_u}

\icmltitlerunning{Efficient Learning with Partially Observed Attributes}

\begin{document}

\twocolumn[
\icmltitle{Efficient Learning with Partially Observed Attributes}

\icmlauthor{Nicol\`o  Cesa-Bianchi}{cesa-bianchi@dsi.unimi.it} 
\icmladdress{DSI, Universit\`{a} degli Studi di Milano, Italy}

\icmlauthor{Shai Shalev-Shwartz}{shais@cs.huji.ac.il} \icmladdress{The Hebrew 
University, Jerusalem, Israel} 

\icmlauthor{Ohad Shamir}{ohadsh@cs.huji.ac.il} \icmladdress{The Hebrew 
University, Jerusalem, Israel}

]

\begin{abstract}
We describe and analyze efficient algorithms for learning a linear predictor
from examples when the learner can only view a few attributes of each
training example. This is the case, for instance, in medical research, where
each patient participating in the experiment is only willing to go through a
small number of tests.
Our analysis bounds the number of additional examples sufficient to
compensate for the lack of full information on each training example.
We demonstrate the efficiency of our algorithms by showing that when running on
digit recognition data, they obtain a high prediction accuracy even when the
learner gets to see only four pixels of each image.
\end{abstract}

\section{Introduction}
Suppose we would like to predict if a person has some disease based on medical
tests. Theoretically, we may choose a sample of the population, perform a large
number of medical tests on each person in the sample and learn from this
information. In many situations this is unrealistic, since patients
participating in the experiment are not willing to go through a large number of
medical tests.
The above example motivates the problem studied in this paper, that is learning
when there is a hard constraint on the number of
attributes the learner may view for each training example. 

We propose an efficient algorithm for dealing with this partial information problem,
and bound the number of additional training examples sufficient to compensate
for the lack of full information on each training example.
Roughly speaking, we actively pick which attributes to observe in a randomized
way so as to construct a ``noisy'' version of \emph{all} attributes.
Intuitively, we can still learn despite the error of this estimate because
instead of receiving the exact value of each individual example
in a small set it suffices to get noisy estimations of many examples.

\subsection{Related Work}
Many methods have been proposed for dealing with missing or partial
information. Most of the approaches do not come with formal guarantees on the
risk of the resulting algorithm, and are not guaranteed to converge in 
polynomial time. The difficulty stems from the exponential number of ways to
complete the missing information. In the framework of generative models, a
popular approach is the Expectation-Maximization (EM) procedure
\citep{DempsterLaRu77}.  The main drawback of the EM approach is that it might
find sub-optimal solutions. In contrast, the methods we propose in this paper
are provably efficient and come with finite sample guarantees on the risk.

Our technique for dealing with missing information borrows ideas from
algorithms for the adversarial multi-armed bandit problem
\citep{AuerCeFrSc03,CesaBianchiLu06}.
Our learning algorithms actively choose which attributes to observe for each
example.  This and similar protocols were studied in the context of active learning
\citep{CohnAtLa94,BalcanBeLa06,Hanneke07a,Hanneke09,BeygelzimerDaLa09}, where
the learner can ask for the target associated with specific examples.

The specific learning task we consider in the paper was first proposed
in \citep{Ben-DavidDi98}, where it is called ``learning with restricted
focus of attention''. \citet{Ben-DavidDi98} considered the
classification setting and showed learnability of several hypothesis
classes in this model, like $k$-DNF and axis-aligned
rectangles. However, to the best of our knowledge, no efficient
algorithm for the class of linear predictors has been proposed.\footnote{
\citet{Ben-DavidDi98} do describe learnability results for similar
classes but only under the restricted family of product
distributions.
} 

A related setting, called budgeted learning, was recently studied - see for
example \cite{DengBoScSuZh07,KapoorGr05} and the references therein. In
budgeted learning, the learner purchases attributes at some fixed cost subject
to an overall budget. Besides lacking formal guarantees, this setting is
different from the one we consider in this paper, because we impose a budget
constraint on the number of attributes that can be obtained for \emph{every}
individual example, as opposed to a global budget. In some applications, such
as the medical application discussed previously, our constraint leads to a more
realistic data acquisition process - the global budget allows to ask for many
attributes of some individual patients while our protocol guarantees a constant
number of medical tests to all the patients. 


Our technique is reminiscent of methods used in the compressed learning
framework \cite{CalderbankJaSc09,ZhouLaWa09}, where data is accessed via a
small set of random linear measurements. Unlike compressed learning, where
learners are both trained and evaluated in the compressed domain, our
techniques are mainly designed for a scenario in which only the access to
training data is restricted.


The ``opposite'' setting, in which full information is given at
training time and the goal is to train a predictor that
depends only on
a small number of attributes at test time, was studied in the context of
learning sparse predictors - see for example \cite{Tibshirani96b} and the wide
literature on sparsity properties of $\ell_1$ regularization. Since
our algorithms also enforce low $\ell_1$ norm, many of those results
can be combined with our techniques to yield an algorithm that views
only $O(1)$ attributes at training time, and 
a number of attributes comparable to the achievable sparsity at test
time. Since our focus in this work is on constrained information at training
time, we do not elaborate on this subject. Furthermore, in some real-world
situations, it is reasonable to assume that attributes are very expensive at
training time but are more easy to obtain at test time. Returning to the
example of medical applications, it is unrealistic to convince patients to
participate in a medical experiment in which they need to go through a lot of
medical tests, but once the system is trained, at testing time, patients who
need the prediction of the system will agree to perform as many medical tests
as needed.

A variant of the above setting is the one
studied by \citet{GreinerGrRo02}, where the learner has all the information
at training time and at test time he tries to actively choose a small
amount of attributes to form a prediction. 
Note that active learning at training time, as we do here, may give more
learning power than active learning at testing time. For example,
we formally prove that while
it is possible to learn a consistent predictor accessing at most
$2$ attributes of each example at training time, it is not possible
(even with an infinite amount of training examples) to build an active
classifier that uses at most $2$ attributes of each example at test time,
and whose error will be smaller than a constant.

\section{Main Results}
In this section we outline the main results. We start with a formal description
of the learning problem.  In linear regression each example is an
instance-target pair, $(\x,y) \in \reals^d \times \reals$. We refer to $\x$ as
a vector of attributes and the goal of the learner is to find a linear
predictor $\x \mapsto \inner{\w,\x}$, where we refer to $\w \in \reals^d$ as
the predictor. The performance of a predictor $\w$ on an instance-target pair,
$(\x,y) \in \reals^d \times \reals$, is measured by a loss function
$\ell(\inner{\w,\x},y)$. For simplicity, we focus on the squared loss function,
$\ell(a,b) = (a-b)^2$, and briefly discuss other loss functions in
\secref{sec:discussion}.
Following the standard framework of statistical learning
\citep{Haussler92,DevroyeGyLu96,Vapnik98}, we model the environment as a joint
distribution $\D$ over the set of instance-target pairs, $\reals^d \times
\reals$. The goal of the learner is to find a predictor with low
risk, defined as the expected loss: $ L_\D(\w) ~\eqdef~ \E_{(\x,y) \sim
  \D}[\ell(\inner{\w,\x},y)] $.  Since the distribution $\D$ is unknown to the
learner he learns by relying on a training set of $m$ examples $S =
(\x_1,y_1),\ldots,(\x_m,y_m)$, which are assumed to be sampled i.i.d. from
$\D$. We denote the training loss by $L_S(\w) \eqdef \frac{1}{m} \sum_{i=1}^m
(\inner{\w,\x_i}-y_i)^2$. We now distinguish between two scenarios:
\begin{itemize}
\item 
{\bf Full information:} The learner receives the entire training
  set. This is the traditional linear regression setting.
\item 
{\bf Partial information:} For each individual example, $(\x_i,y_i)$, the
  learner receives the target $y_i$ but is only allowed to see $k$ attributes
  of $\x_i$, where $k$ is a parameter of the problem. The learner has the
  freedom to actively choose \emph{which} of the attributes will be revealed,
  as long as at most $k$ of them will be given.
\end{itemize}
While the full information case was extensively studied, the partial 
information case is more challenging.  Our approach for dealing with the 
problem of partial information is to rely on algorithms for the full 
information case and to fill in the missing information in a randomized, data 
and algorithmic dependent, way. As a simple baseline, we begin by describing a 
straightforward adaptation of Lasso \citep{Tibshirani96b},
based on a direct nonadaptive estimate of the loss function.
We then turn to describe a more effective approach, which combines a stochastic
gradient descent algorithm called Pegasos \citep{ShalevSiSr07} with the active
sampling of attributes in order to estimate the gradient of the loss at each step.

\subsection{Baseline Algorithm} 
A popular approach for learning a linear regressor is to minimize the empirical
loss on the training set plus a regularization term taking the form of a norm
of the predictor. For example, in ridge regression the regularization term is
$\|\w\|_2^2$ and in Lasso the regularization term is $\|\w\|_1$. Instead of
regularization, we can include a constraint of the form $\|\w\|_1 \le B$ or
$\|\w\|_2 \le B$.  With an adequate tuning of parameters, the regularization
form is equivalent to the constraint form.  In the constraint form, the
predictor is a solution to the following optimization problem:
\begin{equation}\label{eqn:RegLoss}
\min_{\w \in \reals^d} ~ \tfrac{1}{|S|} \sum_{(\x,y) \in S} (\inner{\w,\x}-y)^2
~~\textrm{s.t.}~~ \|\w\|_{p} \le B ~,
\end{equation}
where $S = \{(\x_1,y_1),\ldots,(\x_m,y_m)\}$ is a training set of examples, $B$
is a regularization parameter, and $p$ is $1$ for Lasso and $2$ for ridge
regression.  Standard risk bounds for Lasso imply that if $\hat{\w}$ is a
minimizer of~\eqref{eqn:RegLoss} (with $p=1$), then with probability greater
than $1-\delta$ over the choice of a training set of size $m$ we have
\begin{equation}\label{eqn:LassoBound}
L_\D(\hat{\w}) \le \min_{\w: \|\w\|_1 \le B} L_D(\w) + O\left(\!\!B^2 \,
\sqrt{\frac{\ln(d/\delta)}{m}}\right) ~.
\end{equation}
To adapt Lasso to the partial information case, we first rewrite the squared
loss as follows:
$$
(\inner{\w,\x}-y)^2 = \w^T (\x \x^T) \w -2 y \x^T \w + y^2 ~,
$$ 
where $\w,\x$ are column vectors and $\w^T,\x^T$ are their corresponding
transpose (i.e., row vectors).  Next, we estimate the matrix $\x\x^T$ and the
vector $\x$ using the partial information we have, and then we solve the
optimization problem given in \eqref{eqn:RegLoss} with the estimated values of
$\x \x^T$ and $\x$. To estimate the vector $\x$ we can pick an index $i$
uniformly at random from $[d] = \{1,\ldots,d\}$ and define the estimation to be
a vector $\v$ such that
\begin{equation} \label{eqn:tildexdef}
 {v}_r ~=~ 
\begin{cases}
    d\, x_r & \textrm{if}~r =i \\
    0 & \textrm{else} \end{cases} ~~~.
\end{equation}
It is easy to verify that $\v$ is an unbiased estimate of $\x$, namely, $\E[\v]
= \x$ where expectation is with respect to the choice of the index $i$. When we
are allowed to see $k > 1$ attributes, we simply repeat the above process
(without replacement) and set $\v$ to be the averaged vector.  To estimate the
matrix $\x \x^T$ we could pick two indices $i,j$ independently and uniformly at
random from $[d]$, and define the estimation to be a matrix with all zeros
except $d^2\,x_ix_j$ in the $(i,j)$ entry. However, this yields a non-symmetric
matrix which will make our optimization problem with the estimated matrix
non-convex. To overcome this obstacle, we symmetrize the matrix by adding its
transpose and dividing by $2$. The resulting baseline procedure\footnote{We
  note that an even simpler approach is to arbitrarily assume that the
  correlation matrix is the identity matrix and then the solution to the loss
  minimization problem is simply the averaged vector, $\w = \sum_{(\x,y) \in S}
  y \,\x$. In that case, we can simply replace $\x$ by its estimated vector as
  defined in \eqref{eqn:tildexdef}.  While this naive approach can work on very
  simple classification tasks, it will perform poorly on realistic data sets,
  in which the correlation matrix is not likely to be identity. Indeed, in our
  experiments with the MNIST data set, we found out that this approach
  performed poorly relatively to the algorithms proposed in this paper.} is
given in Algorithm \ref{algo:Baseline}.
\begin{Algorithm}
  \caption{Baseline$(S,k)$\\ 
$S$ --- full information training set with $m$ examples \\
$k$ --- Can view only $k$ elements of each instance in $S$ \\
Parameter: $B$} \label{algo:Baseline}
      \begin{Balgorithm}
      \textsc{Initialize:} $\bar{A} = \gb{0} \in \reals^{d,d}$ ~;~ $\bar{\v} = \gb{0} \in \reals^d$ ~;~
      $\bar{y} = 0$ \\
      {\bf for each} $(\x,y) \in S $ \+ \\
         $\v = \gb{0} \in \reals^d$ \\
         ${A} = \gb{0} \in \reals^{d,d}$ \\
         Choose $C$ uniformly at random from \\
         ~~all subsets of $[d]\times[d]$ of size $k/2$ \\
         {\bf for each} $(i,j) \in C$ \+ \\
         $v_{i} = v_{i} + (d/k)\, x_{i}$ \\
          $v_{j} = v_{j} + (d/k)\, x_{j}$ \\
          ${A}_{i,j} = {A}_{i,j} + (d^2/k)\, x_{i} x_{j}$  \\
          ${A}_{j,i} = {A}_{j,i} + (d^2/k)\, x_{i} x_{j}$ \- \\
         {\bf end }  \\
         $\bar{A} = \bar{A} +  A/m$ \\
         $\bar{\v} = \bar{\v} + 2\,y\,\v/m$ \\
         $\bar{y} = \bar{y} + y^2 / m$ \-\\
      {\bf end } \\
      Let $\tilde{L}_S(\w) = \w^T \bar{A} \w + \w^T\,\bar{\v} + \bar{y}$ \\
      \textsc{Output:} solution of $ \displaystyle 
\min_{\w:\|\w\|_1 \le B} \tilde{L}_S(\w) $
     \end{Balgorithm}
\end{Algorithm}

The following theorem shows that similar to Lasso, the Baseline algorithm is
competitive with the optimal linear predictor with a bounded $L_1$ norm.
\begin{theorem} \label{thm:Baseline}
Let $\D$ be a distribution such that $\prob[\x \in [-1,+1]^d \land y \in
  [-1,+1]] = 1$. Let $\hat{\w}$ be the output of Baseline(S,k), where $|S| =
m$. Then, with probability of at least $1-\delta$ over the choice of the
training set and the algorithm's own randomization we have
$$
L_\D(\hat{\w}) \le \min_{\w: \|\w\|_1 \le B} L_D(\w) + O\left( \frac{(d\,B)^2}{k}\,\sqrt{\frac{\ln(d/\delta)}{m}}\right) ~.
$$
\end{theorem}
The above theorem tells us that for a sufficiently large training set we can
find a very good predictor. Put another way, a large number of examples can
compensate for the lack of full information on each individual example. In
particular, to overcome the extra factor $d^2/k$ in the bound, which does not
appear in the full information bound given in \eqref{eqn:LassoBound}, we need
to increase $m$ by a factor of $d^4/k^2$.

Note that when $k=d$ we do not recover the full information bound. This is
because we try to estimate a matrix with $d^2$ entries using only $k=d <
d^2$ samples. In the next subsection, we describe a better, adaptive
procedure for the partial information case.

\subsection{Gradient-based Attribute Efficient Regression}
In this section, by avoiding the estimation of the matrix $\x \x^T$, we significantly
decrease the number of additional examples sufficient for learning with $k$ attributes
per training example. To do so, we do not try to estimate the loss function but rather estimate the
\emph{gradient} $\nabla \ell(\w) ~=~ 2\,(\inner{\w,\x}-y)\,\x $, with respect
to $\w$, of the squared loss function $(\inner{\w,\x}-y)^2$.  Each 
vector $\w$ can define a probability distribution over $[d]$ by letting 
$\prob[i] = |w_i|/\|\w\|_1$.  We can estimate the gradient using $2$ attributes 
as follows. First, we randomly pick $j$ from $[d]$ according to the 
distribution defined by $\w$. Using $j$ we estimate the term $\inner{\w,\x}$ by 
$\sgn(w_j)\,\|\w\|_1\,x_j$.  It is easy to verify that the expectation of the 
estimate equals $\inner{\w,\x}$. Second, we randomly pick $i$ from $[d]$ 
according to the uniform distribution over $[d]$. Based on $i$, we estimate the 
vector $\x$ as in \eqref{eqn:tildexdef}. Overall, we obtain the following 
unbiased estimation of the gradient:
\begin{equation} \label{eqn:gradientEst}
\tilde{\nabla} \ell(\w) ~=~ 
 2\,(\sgn(w_j)\,\|\w\|_1\,x_j -y)\,\v ~,
\end{equation}
where $\v$ is as defined in \eqref{eqn:tildexdef}.

The advantage of the above approach over the loss based approach we took before
is that the magnitude of each element of the gradient estimate is order of
$d\,\|\w\|_1$. This is in contrast to what we had for the loss based approach,
where the magnitude of each element of the matrix $A$ was order of $d^2$. In
many situations, the $L_1$ norm of a good predictor is significantly smaller
than $d$ and in these cases the gradient based estimate is better than the loss
based estimate.  However, while in the previous approach our estimation did not
depend on a specific $\w$, now the estimation depends on $\w$. We therefore
need an iterative learning method in which at each iteration we 
use the gradient of the loss function on an individual example. Luckily,
the stochastic gradient descent approach conveniently fits our needs.

Concretely, below we describe a variant of the Pegasos algorithm
\citep{ShalevSiSr07} for learning linear regressors. Pegasos tries to minimize
the regularized risk
\begin{equation} \label{eqn:GenObj}
\min_\w ~~\E_{(\x,y) \sim \D}\left[ (\inner{\w,\x}-y)^2 \right] + \lambda \|\w\|_2^2 ~.
\end{equation}
Of course, the distribution $\D$ is unknown, and therefore we cannot hope to
solve the above problem exactly. Instead, Pegasos finds a sequence of weight
vectors that (on average) converge to the solution of \eqref{eqn:GenObj}. We
start with the all zeros vector $\w = \gb{0} \in \reals^d$. Then, at each
iteration Pegasos picks the next example in the training set (which is
equivalent to sampling a fresh example according to $\D$) and calculates the
gradient of the loss function on this example with respect to the current
weight vector $\w$. In our case, the gradient is simply $2 (\inner{\w,\x}-y)
\x$. We denote this gradient vector by $\nabla$. Finally, Pegasos updates the
predictor according to the rule: $\w = (1 - \tfrac{1}{t})\,\w -
\frac{1}{\lambda\,t} \,\nabla$, where $t$ is the current iteration number.

To apply Pegasos in the partial information case we could simply replace the
gradient vector $\nabla$ with its estimation given in
\eqref{eqn:gradientEst}. However, our analysis shows that it is desirable to
maintain an estimation vector $\tilde{\nabla}$ with small magnitude. Since the
magnitude of $\tilde{\nabla}$ is order of $d\,\|\w\|_1$, where $\w$ is the
current weight vector maintained by the algorithm, we would like to ensure that
$\|\w\|_1$ is always smaller than some threshold $B$. We achieve this goal by
adding an additional projection step at the end of each Pegasos's
iteration. Formally, after performing the update we set
\begin{equation} \label{eqn:projection}
\w \leftarrow \argmin_{\u : \|\u\|_1 \le B} \|\u-\w\|_2 ~.
\end{equation}
This projection step can be performed efficiently in time $O(d)$ using the
technique described in \cite{DuchiShSi08}.  A pseudo-code of the resulting
\textbf{A}ttribute \textbf{E}fficient \textbf{R}egression
algorithm is given in Algorithm \ref{algo:AER}.

\begin{Algorithm}
  \caption{AER$(S,k)$\\ 
$S$ --- Full information training set with $m$ examples \\
$k$ --- Access only $k$ elements of each instance in $S$ \\
Parameters: $\lambda,B$} \label{algo:AER}
      \begin{Balgorithm}
      $\w = (0,\ldots,0)$ ~;~ $\bar{\w} = \w$ ~;~ $t=1$ \\
      {\bf for each} $(\x,y) \in S $ \+ \\
         $\v = \gb{0} \in \reals^d$ \\
        Choose $C$ uniformly at random from \\
         ~~all subsets of $[d]$ of size $k/2$ \\
         {\bf for each} $j \in C$ \+ \\
         $v_{j} = v_{j} + \tfrac{2}{k}\,d\, x_{j}$ \-\\
         {\bf end }  \\
         $\hat{y} = 0 $ \\
         {\bf for} $r = 1,\ldots,k/2$ \+ \\
         sample $i$ from $[d]$ based on $\prob[i]=|w_i|/\|\w\|_1$ \\
         $\hat{y} = \hat{y} + \tfrac{2}{k}\,\sgn(w_i) \,\|\w\|_1\, x_{j}$  \-\\
         {\bf end} \\
         $\w = (1-\tfrac{1}{t}) \w - \tfrac{2}{\lambda t}(\hat{y}-y)\v $ \\
         $\w = \argmin_{\u: \|\u\|_1 \le B} \|\u-\w\|_2$ \\
         $\bar{\w} = \bar{\w} + \w / m$ \\
         $t = t+1$\-\\
      {\bf end } \\
      \textsc{Output:} $\bar{\w}$ 
     \end{Balgorithm}
\end{Algorithm}

The following theorem provides convergence guarantees for AER.
\begin{theorem} \label{thm:AER}
Let $\D$ be a distribution such that $\prob[\x \in [-1,+1]^d \land y \in
  [-1,+1]] = 1$.  Let $\w^\star$ be any vector such that
  $\|\w^\star\|_1 \le B$
and $\|\w^\star\|_2 \le B_2$ Then,
\begin{equation*}
\E[L_\D(\bar{\w})] ~\le~ L_\D(\w^\star) + O\left(\frac{d \, (B+1) \,
  B_2}{\sqrt{k}}\,\sqrt{\frac{\ln(m)}{m}}\right) ~,
\end{equation*}
where $|S| = m$, $\bar{\w}$ is the output of AER$(S,k)$ run with $\lambda = ((B+1)
d/B_2)\,\sqrt{\log(m)/(mk)}$, and the expectation is over the choice of $S$ and
over the algorithm's own randomization.
\end{theorem}
For simplicity and readability, in the above theorem we only bounded the
expected risk.  It is possible to obtain similar guarantees with high
probability by relying on Azuma's inequality ---see for example
\cite{CesaBianchiCoGe04}.

Note that $\|\w^\star\|_2 \le \|\w^\star\|_1 \le B$, so
\thmref{thm:AER} implies that
$$ 
L_\D(\bar{\w}) \le \min_{\w: \|\w\|_1 \le B} L_D(\w) + O\left(
\frac{d\,B^2}{\sqrt{k}}\,\sqrt{\frac{\ln(m)}{m}}\right) ~.
$$ 
Therefore, the bound for AER is much better\footnote{When comparing bounds,
  we ignore logarithmic terms. Also, in this discussion we assume that
$B_1$ and $B_2$ are at least $1$. } than the bound for Baseline: instead of
$d^2/k$ we have $d/\sqrt{k}$.

It is interesting to compare the bound for AER to the Lasso bound in the full
information case given in~\eqref{eqn:LassoBound}. As it can be seen, to achieve
the same level of risk, AER needs a factor of $d^2/k$ more examples than
the full information Lasso.\footnote{We note that when $d=k$ we still do not
  recover the full information bound. However, it is possible to improve the
  analysis and replace the factor $d/\sqrt{k}$ with a factor $d
  \max_t\|\x_t\|_2/k$. }  Since
each AER example uses only $k$ attributes while each Lasso example uses all
$d$ attributes, the ratio between the total number of \emph{attributes} AER
needs and the number of attributes Lasso needs to achieve the same error is
$O(d)$. Intuitively, when having $d$ times total number of attributes, we can
fully compensate for the partial information protocol.

However, in some situations even this extra $d$ factor is not needed.
Suppose we know that the vector $\w^\star$, which minimizes the risk, is
dense. That is, it satisfies $\|\w^\star\|_1 \approx \sqrt{d}\,\|\w^\star\|_2$.
In this case, choosing $B_2 = B/\sqrt{d}$, the bound in \thmref{thm:AER}
becomes order of $B^2 \sqrt{d/k} \sqrt{1/m}$.  Therefore, the number of
examples AER needs in order to achieve the same error as Lasso is only
a factor $d/k$ more than the number of examples Lasso uses. But, this implies
that both AER and Lasso needs the same number of \emph{attributes} in
order to achieve the same level of error! Crucially, the above holds only if
$\w^\star$ is dense. When $\w^\star$ is sparse we have $\|\w^\star\|_1
\approx \|\w^\star\|_2$ and then AER needs more attributes than Lasso.

One might wonder whether a more clever active sampling strategy could attain
in the sparse case the performance of Lasso while using the same number of
attributes. The next subsection shows that this is not possible in general.

\subsection{Lower bounds and negative results} \label{sec:lower}
We now show (proof in the appendix) that any attribute efficient algorithm
needs in general order of $d/\epsilon$ examples for learning an
$\epsilon$-accurate sparse linear predictor.  Recall that the upper bound of
AER implies that order of $d^2 (B+1)^2 B_2^2 / \epsilon^2$ examples are
sufficient for learning a predictor with $L_\D(\w) - L_\D(\w^\star) <
\epsilon$.  Specializing this sample complexity bound of AER to the $\w^\star$
described in \thmref{thm:lower} below, yields that $O(d^2 /\epsilon)$ examples
are sufficient for AER for learning a good predictor in this case. That is, we
have a gap of factor $d$ between the lower bound and the upper bound, and it
remains open to bridge this gap.
\begin{theorem} \label{thm:lower}
For any $\epsilon \in (0,1/16)$, $k$, and $d \ge 4k$, there exists
a distribution over examples and a weight vector $\w^\star$, with
$\|\w^\star\|_0 = 1$ and $\|\w^\star\|_2=\|\w^\star\|_1 = 2\sqrt{\epsilon}$,
such that any attribute efficient regression algorithm accessing at most
$k$
attributes per training example must see (in expectation) at least
$\Omega\left(\tfrac{d}{k \epsilon}\right)$ examples in order to learn a linear predictor $\w$
with $L_\D(\w) - L_\D(\w^\star) < \epsilon$.
\end{theorem}
Recall that in our setting, while at training time the learner can only view
$k$ attributes of each example, at test time all attributes can
be observed.
The setting of \citet{GreinerGrRo02}, instead, assumes that at test time the learner
cannot observe all the attributes. The following theorem shows that if 
a learner can view at most $2$ attributes at test time then it is
impossible to give accurate predictions at test time even when
the optimal linear predictor is known.
\begin{theorem} \label{thm:lowertest}
There exists a weight vector $\w^\star$ and a distribution $\D$ such
that $L_\D(\w^{\star}) = 0$ while any algorithm $A$ that gives predictions
$A(\x)$ while viewing only $2$ attributes of each $\x$ must have
$L_\D(A) \ge 1/9$. 
\end{theorem}
The proof is given in the appendix.
This negative result highlights an interesting phenomenon.
We can learn an arbitrarily accurate predictor $\w$
from partially observed examples. However, even if we know the optimal
$\w^\star$, we might not be able to accurately predict a new
partially observed example.

\section{Proof Sketch of \thmref{thm:AER}} \label{sec:sketch} 

Here we only sketch the proof of \thmref{thm:AER}. A complete proof of all our
theorems is given in the appendix.

We start with a general logarithmic regret bound for strongly convex
functions \cite{HazanKaKaAg06,KakadeSh08}. The regret bound implies
the following. Let $\z_1,\ldots,\z_m$ be a sequence of vectors, each
of which has norm bounded by $G$. Let $\lambda > 0$ and consider the
sequence of functions $g_1,\ldots,g_m$ such that $g_t(\w) =
\tfrac{\lambda}{2} \|\w\|^2 + \inner{\z_t,\w}$. Each $g_t$ is
$\lambda$-strongly convex (meaning, it is not too flat), and
therefore  regret bounds for strongly convex functions tell us that
there is a way to construct a sequence of vectors $\w_1,\ldots,\w_m$
such that for any $\w^\star$ that satisfies $\|\w^\star\|_1 \le B$ we have
\[
\frac{1}{t} \sum_{t=1}^m g_t(\w_t) - \frac{1}{t} \sum_{t=1}^m
g_t(\w^\star) \le O\left(\tfrac{G^2\,\log(m)}{\lambda\,m}\right) ~. 
\]
With an appropriate choice of $\lambda$, and with the assumption
$\|\w^\star\|_2 \le B_2$, the above inequality implies that $ \tfrac{1}{m}
\sum_{t=1}^m \inner{\z_t,\w_t-\w^\star} \le \alpha $ where $\alpha =
O\left(\tfrac{G\,B_2\,\log(m)}{\sqrt{m}}\right) $.  This holds for any sequence
of $\z_1,\ldots,\z_m$, and in particular, we can set $\z_t = 2(\hat{y}_t - y_t)
\v_t$. Note that $\z_t$ is a random vector that depends both on the value of
$\w_t$ and on the random bits chosen on round $t$. Taking conditional
expectation of $\z_t$ w.r.t. the random bits chosen on round $t$ we obtain that
$\E[\z_t|\w_t]$ is exactly the gradient of $(\inner{\w,\x_t}-y_t)^2$ at $\w_t$,
which we denote by $\nabla_t$. From the convexity of the squared loss, we can
lower bound $\inner{\nabla_t,\w_t-\w^\star}$ by
$(\inner{\w_t,\x_t}-y_t)^2-(\inner{\w^\star,\x_t}-y_t)^2$. That is, in
expectation we have that
\[
\E \left[ \tfrac{1}{m} \sum_{t=1}^m \left(
    (\inner{\w_t,\x_t}-y_t)^2-(\inner{\w^\star,\x_t}-y_t)^2\right)
\right]
\le \alpha ~.
\]
Taking expectation w.r.t. the random choice of the examples from $\D$, 
denoting $\bar{\w} = \tfrac{1}{m} \sum_{t=1}^m$, and using Jensen's
inequality we get that
$
\E[ L_\D(\bar{\w})] \le L_\D(\w^\star) + \alpha $.
Finally, we need to make sure that $\alpha$ is not too large. The only
potential danger is that $G$, the bound on the norms of
$\z_1,\ldots,\z_m$, will be large. We make sure this cannot happen by
restricting each $\w_t$ to the $\ell_1$ ball of radius $B$, which
ensures that $\|\z_t\| \le O((B+1)d)$ for all $t$.

\section{Experiments} \label{sec:experiments}
We performed some preliminary experiments to test the behavior of our algorithm
on the well-known MNIST digit recognition dataset \cite{LeCunBoBeHa98},
which contains 70,000 images ($28 \times 28$ pixels each) of the digits $0-9$.
The advantages of this dataset for our purposes is that it is not a small-scale
dataset, has a reasonable dimensionality-to-data-size ratio, and the setting is
clearly interpretable graphically. While this dataset is designed for
classification (e.g.  recognizing the digit in the image), we can still apply
our algorithms on it by regressing to the label. 

\begin{figure}[t]
\begin{center}
\includegraphics[trim = 40mm 14mm 20mm 10mm,clip=true,scale=0.35]{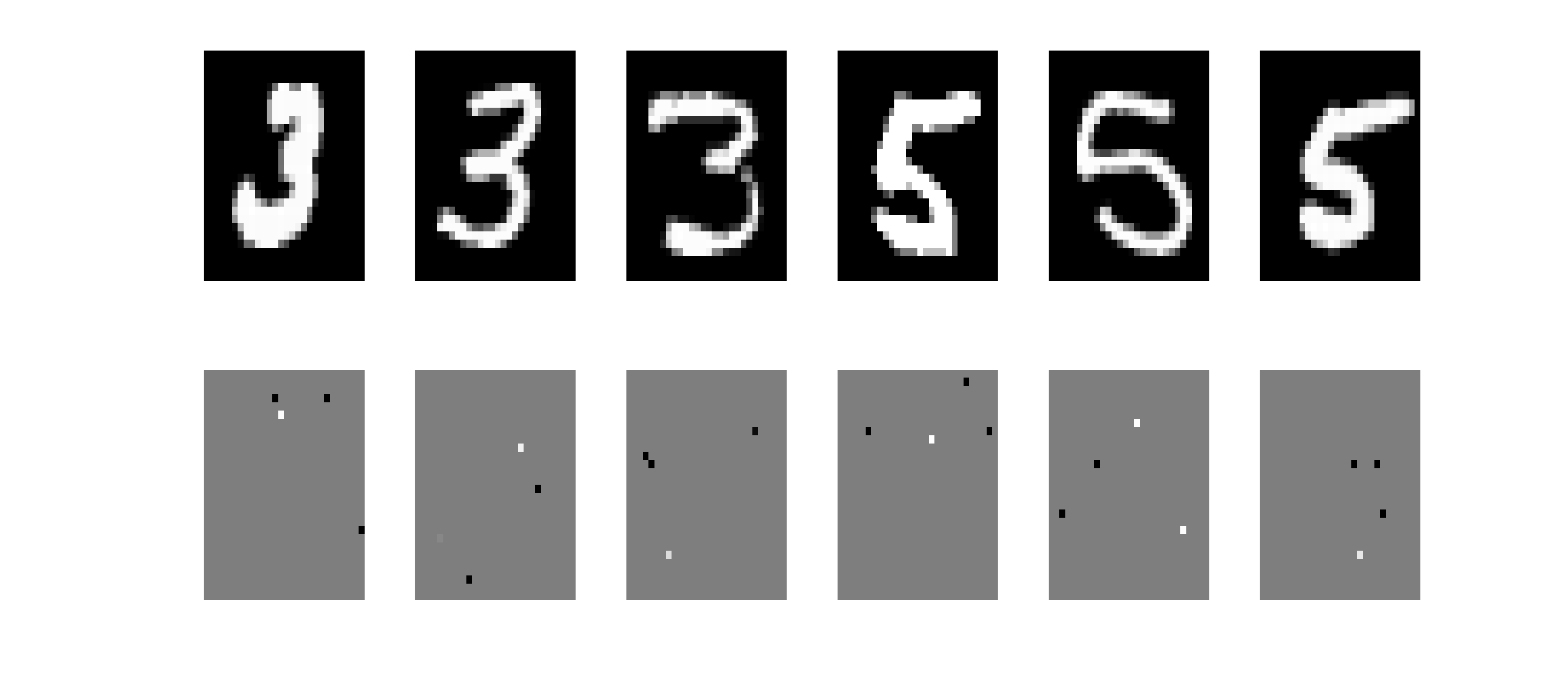}
\end{center}
\caption{\small In the upper row six examples from the training set (of
  digits 3 and 5) are shown. In the lower row we show the same six examples,
  where only four randomly sampled pixels from each original image are displayed.}
\label{fig:examples}
\vspace{-0.5cm}
\end{figure}

First, to demonstrate the hardness of our settings, we provide in
\figref{fig:examples} below some examples of images from the dataset, in the
full information setting and the partial information setting. The upper row
contains six images from the dataset, as available to a full-information
algorithm. A partial-information algorithm, however, will have a much more
limited access to these images. In particular, if the algorithm may only choose
$k=4$ pixels from each image, the same six images as available to it might look
like the bottom row of \figref{fig:examples}. 

We began by looking at a dataset composed of ``$3$ vs.\ $5$'', where all the $3$ 
digits were labeled as $-1$ and all the $5$ digits were labeled as $+1$. We 
ran four different algorithms on this dataset: the simple Baseline algorithm, 
AER, as well as ridge regression and Lasso for comparison (for Lasso, we 
solved \eqref{eqn:RegLoss} with $p=1$). Both ridge regression and Lasso were 
run in the full information setting: Namely, they enjoyed full access to all 
attributes of all examples in the training set. The Baseline algorithm and 
AER, however, were given access to only $4$ attributes from each training 
example.

\begin{figure}[ht]
\begin{center}
\includegraphics[trim = 0mm 5mm 0mm 12mm,clip=true, scale=0.7]{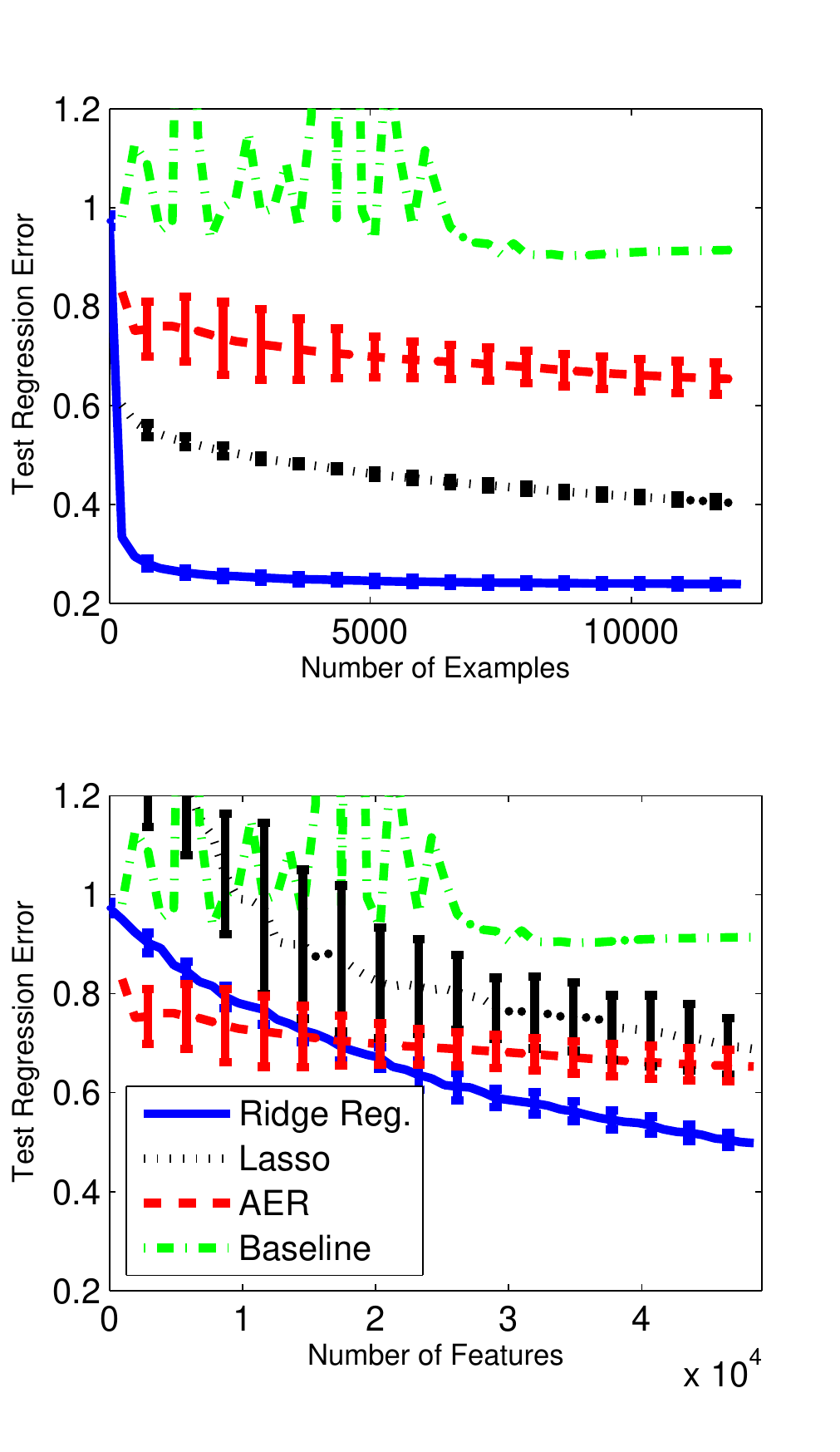} \vspace{-0.5cm}
\end{center}
\caption{\small Test regression error for each of the $4$ algorithms, over
  increasing prefixes of the training set for ``3 vs.\ 5''. The results are
  averaged over $10$ runs.}
\vspace{-0.3cm}
\label{fig:mnist_35}
\end{figure}

We randomly split the dataset into a training set and a test set (with the 
test set being $10\%$ of the original dataset). For each algorithm, parameter 
tuning was performed using $10$-fold cross validation. Then, we ran the 
algorithm on increasingly long prefixes of the training set, and measured the 
average regression error $(\inner{\w,\x}-y)^2$ on the test set. The results 
(averaged over runs on $10$ random train-test splits) are presented in 
\figref{fig:mnist_35}. In the upper plot, we see how the test regression error 
improves with the number of examples. The Baseline algorithm is highly 
unstable at the beginning, probably due to the ill-conditioning of the 
estimated covariance matrix, although it eventually stabilizes (to prevent a 
graphical mess at the left hand side of the figure, we removed the error bars 
from the corresponding plot). Its performance is worse than AER, completely 
in line with our earlier theoretical analysis.

The bottom plot of \figref{fig:mnist_35} is similar, only that now the 
$X$-axis represents the accumulative number of attributes seen by each 
algorithm rather than the number of examples. For the partial-information 
algorithm, the graph ends at approximately 49,000 attributes, which is the 
total number of attributes accessed by the algorithm after running over all 
training examples, seeing $k=4$ pixels from each example. However, for the 
full-information algorithm 49,000 attributes are already seen after just $62$ 
examples. When we compare the algorithms in this way, we see that our AER 
algorithm achieves excellent performance for a given attribute budget, 
significantly better than the other $L_1$-based algorithms, and even 
comparable to full-information ridge regression.

Finally, we tested the algorithms over $45$ datasets generated from MNIST, one 
for each possible pair of digits. For each dataset and each of $10$ random 
train-test splits, we performed parameter tuning for each algorithm 
separately, and checked the average squared error on the test set. 
The median test errors over all datasets  
are presented in the table below.
\begin{center}
\begin{tabular}{|c|c|c|}
\hline 
& & Test Error \\ \hline
Full Information & \textbf{Ridge} & 0.110 \\
& \textbf{Lasso} & 0.222 \\\hline
Partial Information & \textbf{AER} & 0.320 \\
&\textbf{Baseline}  & 0.815 \\
\hline
\end{tabular}
\end{center}
      
As can be seen, the AER algorithm manages to achieve good
performance, not much worse than the full-information Lasso algorithm. The
Baseline algorithm, however, achieves a substantially worse performance, in 
line with our theoretical analysis above. 
We also calculated the test classification error of AER,
i.e. $\text{sign}(\inner{\w,\x})\neq y$, and found out that AER, 
which can see only $4$ pixels per image, 
usually perform only a little worse than the full-information algorithms 
(ridge regression and Lasso), which enjoy full access to all $784$ pixels in 
each image. In particular, the median test classification errors of AER,
Lasso, and Ridge are $3.5\%,~1.1\%,$ and $1.3\%$ respectively. 


\section{Discussion and Extensions} \label{sec:discussion}

In this paper, we provided an efficient algorithm for learning when only a few
attributes from each training example can be seen. The algorithm comes with
formal guarantees, is provably competitive with algorithms which enjoy full
access to the data, and seems to perform well in practice. We also presented
sample complexity lower bounds, which are only a factor $d$ smaller than the
upper bound achieved by our algorithm, and it remains open to bridge this gap.

Our approach easily extends to other gradient-based algorithms besides
Pegasos. For example, generalized additive algorithms such as
$p$-norm Perceptrons and Winnow - see, e.g., \cite{CesaBianchiLu06}.

An obvious direction for future research is how to deal with loss functions
other than the squared loss. In upcoming work on a related problem, we develop
a technique which allows us to deal with arbitrary analytic loss functions, but
in the setting of this paper will lead to sample complexity bounds which are
exponential in $d$. Another interesting extension we are considering
is connecting our results to the field of privacy-preserving learning
\citep{Dwork08}, where the goal is to exploit the attribute efficiency
property in order to prevent acquisition of information about individual data
instances.

{\small
\bibliographystyle{plainnat}
\bibliography{bib}
}

\onecolumn
\appendix

\section{Proofs}

\subsection{Proof of \thmref{thm:Baseline}}

To ease our calculations, we first show that sampling $k$ elements without
replacements and then averaging the result has the same expectation as sampling
just once. In the lemma below, for a set $C$ we denote the uniform distribution
over $C$ by $U(C)$.
\begin{lemma} \label{lem:replacements}
Let $C$ be a finite set and let $f: C \to \reals$ be an arbitrary function. 
Let $C_k = \{C' \subset C : |C'|=k\}$. Then,
$$
\E_{C' \sim U(C_k)}[ \tfrac{1}{k} \sum_{c \in C'} f(c)] ~=~ 
\E_{c \sim U(C)}[ f(c)] ~.
$$
\end{lemma}
\begin{proof}
Denote $|C|=n$. We have:
\begin{equation*}
\begin{split}
\E_{C' \sim U(C_k)}[& \tfrac{1}{k} \sum_{c \in C'} f(c)] = \frac{1}{{n \choose k}} \sum_{C' \in C_k} 
\tfrac{1}{k} \sum_{c \in C'} f(c)\\
&= \frac{1}{k{n \choose k}} \sum_{c \in C} f(c) |\{C' \in C_k : c \in C'\}| \\
&= \frac{{n-1 \choose k-1}}{k{n \choose k}} \sum_{c \in C} f(c)  \\
&= \frac{(n-1)! k!(n-k)!}{k n! (k-1)! (n-k)!} \sum_{c \in C} f(c)  \\
&= \frac{1}{n} \sum_{c \in C} f(c) \\
&= \E_{c \sim U(C)}[f(c)] ~.
\end{split}
\end{equation*}
\end{proof}

To prove \thmref{thm:Baseline} we first show that the estimation matrix
constructed by the Baseline algorithm is likely to be close to the true 
correlation matrix over the training set.

\begin{lemma} \label{lem:Aconcentration}
Let $A_t$ be the matrix constructed at iteration $t$ of the Baseline algorithm
and note that $\bar{A} = \frac{1}{m} \sum_{t=1}^m A_t$.  Let $X = \frac{1}{m}
\sum_{t=1}^m \x_t \x_t^T$. Then, with probability of at least $1-\delta$ over
the algorithm's own randomness we have that
$$ 
\forall r,s~~ |\bar{A}_{r,s}-X_{r,s}| ~\le~
\frac{d^2}{k}\,\cdot\,\sqrt{\frac{2\ln(2d^2/\delta)}{m}} ~.
$$
\end{lemma}
\begin{proof}
Based on \lemref{lem:replacements}, it is easy to verify that $\E[A_t] = \x_t^T
\x_t$.  Additionally, since we sample without replacements, each element of
$A_t$ is in $[-d^2/k,d^2/k]$ (because we assume $\|\x_t\|_\infty \le
1$). Therefore, we can apply Hoeffding's inequality on each element of
$\bar{A}$ and obtain that
$$
\prob[|\bar{A}_{r,s} - X_{r,s}| > \epsilon] ~\le~ 2 e^{-m\,k^2\,\epsilon^2/(2d^4)} ~.
$$
Combining the above with the union bound we obtain that
$$
\prob[\exists (r,s) : |\bar{A}_{r,s} - X_{r,s}| > \epsilon] ~\le~ 2 d^2\,e^{-m\,k^2\,\epsilon^2/(2d^4)} ~.
$$ 
Calling the right-hand-side of the above $\delta$ and rearranging terms we
conclude our proof.
\end{proof}

Next, we show that the estimate of the linear part of the objective function is
also likely to be accurate.
\begin{lemma} \label{lem:vconcentration}
Let $\v_t$ be the vector constructed at iteration $t$ of the Baseline algorithm
and note that $\bar{\v} = \frac{1}{m} \sum_{t=1}^m 2 y_t \v_t$. Let $\bar{\x} =
\frac{1}{m} \sum_{t=1}^m 2 y_t \x_t$. Then, with probability of at least
$1-\delta$ over the algorithm's own randomness we have that
$$ \|\bar{\v}-\bar{\x}\|_\infty ~\le~
\frac{d}{k}\,\cdot\,\sqrt{\frac{8\ln(2d/\delta)}{m}} ~.
$$
\end{lemma}
\begin{proof}
Based on \lemref{lem:replacements}, it is easy to verify that
$\E[2 y_t \v_t] = 2 y_t \x_t$.  Additionally, since we sample $k/2$ pairs
without replacements, each element of $\v_t$ is in $[-2d/k,2d/k]$ (because we
assume $\|\x_t\|_\infty \le 1$) and thus each element of $2 y_t \v_t$ is in
$[-4d/k,4d/k]$ (because we assume that $|y_t| \le 1$). Therefore, we can
apply Hoeffding's inequality on each element of $\bar{\v}$ and obtain that $$
\prob[|\bar{v}_r - \bar{x}_r| > \epsilon] ~\le~ 2
e^{-m\,k^2\,\epsilon^2/(8d^2)} ~.  $$ Combining the above with the union
bound we obtain that $$ \prob[\exists (r,s) : |\bar{A}_{r,s} - X_{r,s}| >
\epsilon] ~\le~ 2 \,d\,e^{-m\,k^2\,\epsilon^2/(8d^2)} ~.  $$ Calling the
right-hand-side of the above $\delta$ and rearranging terms we conclude our
proof.  
\end{proof}

We next show that the estimated training loss found by the Baseline algorithm,
$\tilde{L}_S(\w)$, is close to the true training loss.
\begin{lemma} \label{lem:tildeLs}
With probability greater than $1-\delta$ over the Baseline Algorithm's own 
randomization, for all $\w$ such that $\|\w\|_1 \le B$ we have that
$$
|\tilde{L}_S(\w) - L_S(\w)| \le
O\left(\frac{B^2\,d^2}{k}\,\cdot\,\sqrt{\frac{\ln(d/\delta)}{m}}\right) ~.
$$
\end{lemma}
\begin{proof}
Combining \lemref{lem:Aconcentration} with the boundedness of
$\|\w\|_1$ and using Holder's inequality twice we easily get that $$ |\w^T
(\bar{A}-X) \w| \le
\frac{B^2\,d^2}{k}\,\cdot\,\sqrt{\frac{2\ln(2d^2/\delta)}{m}} ~.  $$
Similarly, using \lemref{lem:vconcentration} and Holder's inequality, $$
|\w^T (\bar{\v} - \bar{\x})| \le
\frac{B\,d}{k}\,\cdot\,\sqrt{\frac{8\ln(2d/\delta)}{m}} ~.  $$ Combining the
above inequalities with the union bound and the triangle inequality we
conclude our proof. 
\end{proof}

We are now ready to prove \thmref{thm:Baseline}.  First, using standard risk
bounds (based on Rademacher complexities\footnote{ To bound the Rademacher
  complexity, we use the boundedness of $\|\w\|_1,\|\x\|_\infty,|y|$ to get
  that the squared loss is $O(B)$ Lipschitz on the domain. Combining this with
  the contraction principle yields the desired Rademacher bound.}) we know that
with probability greater than $1-\delta$ over the choice of a training set of
$m$ examples, for all $\w$ s.t. $\|\w\|_1 \le B$, we have that
$$
|L_S(\w)-L_\D(\w)| \le O\left(B^2\,\sqrt{\frac{\ln(d/\delta)}{m}}\right) ~.
$$ 
Combining the above with \lemref{lem:tildeLs} we obtain that for any $\w$
s.t. $\|\w\|_1 \le B$,
\begin{equation*} 
\begin{split}
|L_\D(\w)&-\tilde{L}_S(\w)| \\
&\le~ |L_\D(\w)-L_S(\w)| + |L_S(\w)-\tilde{L}_S(\w)| \\
&\le~  O\left(\frac{B^2\,d^2}{k}\,\cdot\,\sqrt{\frac{\ln(d/\delta)}{m}}\right) ~.
\end{split}
\end{equation*}
The proof of \thmref{thm:Baseline} follows since the Baseline algorithm 
minimizes $\tilde{L}_S(\w)$. 

\subsection{Proof of \thmref{thm:AER}}

We start with the following lemma.
\begin{lemma} \label{lem:regret}
Let $y_t,\hat{y}_t,\v_t,\w_t$ be the values of $y,\hat{y},\v,\w$, respectively,
at iteration $t$ of the AER algorithm.  Then, for any vector $\w^\star$
s.t. $\|\w^\star\|_1 \le B$ we have
\begin{equation*}
\begin{split}
&\sum_{t=1}^m \left(\tfrac{\lambda}{2} \|\w_t\|_2^2 +
  2(\hat{y}_t-y_t)\inner{\v_t,\w_t}\right) ~\le\\ & \sum_{t=1}^m
  \left(\tfrac{\lambda}{2} \|\w^\star\|_2^2 +
  2(\hat{y}_t-y_t)\inner{\v_t,\w^\star}\right) + O\left(\tfrac{((B+1)d)^2/k
    \log(m)}{\lambda}\right) ~.
\end{split}
\end{equation*}
\end{lemma}
\begin{proof}
The proof follows directly from logarithmic regret bounds for strongly convex
functions \citep{HazanKaKaAg06,KakadeSh08} by noting that according to our
construction, $\max_t 2(\hat{y}_t-y_t)\|\v_t\|_2 \le O((B+1)\,d/\sqrt{k})$.
\end{proof}

Let $B_2$ be such that $\|\w^\star\|_2 \le B_2$ and choose $\lambda = ((B+1)
d/B_2)\,\sqrt{\log(m)/(mk)}$.  Since $\lambda\|\w_t\|^2 \ge 0$ we obtain from
\lemref{lem:regret} that
\begin{equation} \label{eqn:regret1}
\begin{split}
&\sum_{t=1}^m 2(\hat{y}_t-y_t)\inner{\v_t,\w_t-\w^\star} \le
  \tfrac{m\lambda\|\w^\star\|_2^2}{2}\\ & + O\left(\tfrac{((B+1)d)^2/k
    \log(m)}{\lambda}\right) = \underbrace{O\left(\tfrac{d}{\sqrt{k}}\, (B+1) \,
    B_2\, \sqrt{\tfrac{\log(m)}{m}}\right)}_{\eqdef \alpha}~.
\end{split}
\end{equation}
For each $t$, let $\nabla_t = 2(\inner{\w_t,\x_t}-y_t)\x_t$ and
$\tilde{\nabla}_t = 2(\hat{y}_t-y_t)\v_t$. Taking expectation of
\eqref{eqn:regret1} with respect to the algorithm's own randomization, and
noting that the conditional expectation of $\tilde{\nabla}_t$ equals
$\nabla_t$, we obtain
\begin{equation} \label{eqn:regret2}
\begin{split}
&\E\left[\sum_{t=1}^m \inner{\nabla_t,\w_t-\w^\star}\right] 
\le \alpha~.
\end{split}
\end{equation}
From the convexity of the squared loss we know that
$$ 
(\inner{\w_t,\x_t}-y_t)^2 - (\inner{\w^\star,\x_t}-y_t)^2 \le
\inner{\nabla_t,\w_t-\w^\star} ~.
$$
Combining with \eqref{eqn:regret2} yields
\begin{equation} \label{eqn:regret3}
\begin{split}
&\E\left[\sum_{t=1}^m (\inner{\w_t,\x_t}-y_t)^2 - (\inner{\w^\star,\x_t}-y_t)^2\right] 
\le \alpha~.
\end{split}
\end{equation}
Taking expectation again, this time with respect to the randomness in choosing
the training set, and using the fact that $\w_t$ only depends on previous
examples in the training set, we obtain that
\begin{equation} \label{eqn:regret4}
\begin{split}
&\E\left[\sum_{t=1}^m L_\D(\w_t) - L_\D(\w^\star)\right] 
\le \alpha ~.
\end{split}
\end{equation}
Finally, from Jensen's inequality we know that $\E[ \tfrac{1}{m} \sum_{t=1}^m
  L_\D(\w_t)] \ge \E[ L_\D(\bar{\w})]$ and this concludes our proof.

\subsection{Proof of \thmref{thm:lower}}
The outline of the proof is as follows. We define a specific
distribution such that only one ``good'' feature is slightly correlated with
the label. We then show that if some algorithm learns a linear
predictor with an extra risk of at most $\epsilon$, then it must know
the value of the 'good' feature. Next, we construct a variant of a multi-armed
bandit problem out of our distribution and show that a good learner can
yield a good prediction strategy.  Finally, we adapt a lower bound for
the multi-armed bandit problem given in \cite{AuerCeFrSc03}, to
conclude that in our case no learner can be too good. 

\paragraph{The distribution:} 
  We generate a joint distribution over $\reals^d \times \reals$ as
  follows. Choose some $j \in [d]$.  First,
  each feature is generated i.i.d. according to $\prob[x_i =
  1]=\prob[x_i=-1]=\thalf$. Next, given $\x$ and $j$, $y$ is
  generated according to $\prob[y =  x_j] = \thalf + p$ and
  $\prob[y = - x_j] = \thalf-p$, where $p$ is set to be
  $\sqrt{\epsilon}$. Denote by $P_j$ the distribution mentioned above assuming the ``good'' feature
is $j$. Also denote by $P_u$ the uniform distribution over $\{\pm
1\}^{d+1}$. Analogously, we denote by $\Ej$ and $\Eu$ expectations
w.r.t. $P_j$ and $P_u$. 

\paragraph{A good regressor ``knows'' $j$}: 
We now show that if we have a good linear regressor than we can know
the value of $j$. 
The optimal linear predictor is $\w^\star = 2 p \e^j$ and
the risk of $\w^\star$ is
$$
L_\D(\w^\star) = \E[(\inner{\w^\star,\x}-y)^2] = \left(\thalf + p\right) (1-2p)^2 
+ \left(\thalf - p\right) (1+2p)^2 = 1 + 4 p^2 - 8p^2 = 1 - 4p^2~.
$$
The risk of an arbitrary weight vector under the aforementioned
distribution is:
\begin{equation}
L_\D(\w) = \E_{\x,y}[(\inner{\w,\x}-y)]^2 ~=~ \sum_{i \neq j} w_i^2 + \E[(w_jx_j
-y)^2] ~=~ \sum_{i \neq j} w_i^2 + w_j^2 + 1 - 4p w_j ~.
\end{equation}
Suppose that $L_\D(\w) - L_\D(\w^\star) < \epsilon$. This implies that:
\begin{enumerate}
\item For all $i \neq j$ we have $w_i^2 < \epsilon$, or equivalently,
  $|w_i| < \sqrt{\epsilon}$.
\item $1 + w_j^2 - 4pw_j - (1-4p^2) < \epsilon$ and thus 
$|w_j-2p| < \sqrt{\epsilon}$ which gives
  $|w_j| > 2p - \sqrt{\epsilon}$ 
\end{enumerate}
Since we set  $p = \sqrt{\epsilon}$, the above implies that we can identify the
value of $j$ from any $\w$ whose risk is strictly smaller than $L_\D(\w^\star)+\epsilon$.

\paragraph{Constructing a variant of a multi-armed bandit problem:}
We now construct a variant of the multi-armed bandit problem out of the
distribution $P_j$. Each $i \in [d]$ is an arm and the reward of pulling $i$ is
$\tfrac{1}{2}|x_i+y|\in\{0,1\}$. Unlike standard multi-armed bandit problems,
here at each round the learner chooses $K$ arms $a_{t,1},\dots,a_{t,K}$, which
correspond to the $K$ atributes accessed at round $t$, and his reward is
defined to be the average of the rewards of the chosen arms.  At the end of
each round the learner observes the value of $\x_t$ at $a_{t,1},\dots,a_{t,K}$,
as well as the value of $y_t$.  Note that the expected reward is $\tfrac{1}{2}
+ p \tfrac{1}{K} \sum_{i=1}^K \indct{a_{t,i}=j}$. Therefore, the total expected
reward of an algorithm that runs for $T$ rounds is upper bounded by
$\tfrac{1}{2}T + p \E[N_j]$, where $N_j$ is the number of times $j \in
\{a_{t,1},\dots,a_{t,K}\}$.

\paragraph{A good learner yields a strategy:}
Suppose that we have a learner that can learn a linear predictor with
$L_\D(\w)-L_\D(\w^\star) < \epsilon$ using
$m$ examples (on average). Since we have shown that once
$L_\D(\w)-L_\D(\w^\star) < \epsilon$ we know the value of $j$, we can
construct a strategy for the multi-armed bandit problem in a
straightforward way; Simply use the first $m$ examples to learn $\w$
and from then on always pull the arm $j$, namely, $a_{t,1}=\ldots=a_{t,K}=j$.
The expected reward of this algorithm is at least
\[
    \tfrac{1}{2}m + (T-m)\left(\tfrac{1}{2}+p\right) = \tfrac{1}{2}T +(T-m)p ~.
\]

\paragraph{An upper bound on the reward of any strategy:}
Consider an arbitrary prediction algorithm. At round $t$ the algorithm
uses the history (and its own random bits, which we can assume are set
in advance) to ask for the current $K$ attributes $a_{t,1},\dots,a_{t,K}$.
The history is the value of $\x_s$ at $a_{s,1},\dots,a_{s,K}$ as well
as the value of $y_s$, for all $s < t$. That is, we can denote the history
at round $t$ to be  $\r^t=(r_{1,1},\dots,r_{1,K+1}),\ldots,(r_{t-1,1},\dots,r_{t-1,K+1})$.
Therefore, on round $t$ the algorithm uses a mapping from $\r^t$ to $[d]^K$.
We use $\r$ as a shorthand for $\r^{T+1}$. The following lemma shows that
any function of the history cannot distinguish too well between the
distribution $P_j$ and the uniform distribution. 

\begin{lemma} \label{lem:A1}
Let $f : \{-1,1\}^{(K+1)T} \to [0,M]$ be any function defined on a history
sequence
$\r=(r_{1,1},\dots,r_{1,K+1}),\ldots,(r_{T,1},\dots,r_{T,K+1})$. Let
$N_j$ be the number of times the algorithm calculating $f$ picks action $j$
among the selected arms. Then,
$$
\Ej[f(\r)] \le \Eu[f(\r)] + M \sqrt{- \log(1-4p^2)\Eu[N_j]
  } ~.
$$
\end{lemma}
\begin{proof}
For any two distributions $P,Q$ we let $\|P-Q\|_1 = \sum_{\r}
|P[\r]-Q[\r]|$ be the total variation distance and let $KL(P,Q) =
\sum_{\r} P[\r]\log(P[\r]/Q[\r])$ be the KL divergence. Using Holder
inequality we know that $\Ej[f(\r)]-\Eu[f(\r)] \le M \|P_j-P_u\|_1$.
Additionally, using Pinsker's inequality we have $\half \|P_j-P_u\|_1^2
\le KL(P_u,P_j)$. Finally, the chain rule and simple calculations yield, 
\[
\begin{split}
KL(P_u,P_j) &= \sum_{\r} \left(\thalf\right)^{(K+1)T} \sum_{t=1}^T
\log\left(  \frac{P_u[r_{t,\cdot}\mid\r^{t-1}]}{P_j[r_{t,\cdot}\mid\r^{t-1}]}\right)
\\
&=  \sum_{\r} \left(\thalf\right)^{(K+1)T} \sum_{t=1}^T
\log\left(\frac{\left(\thalf\right)^{K+1}}{\left(\thalf\right)^{K+1}
+ \left(\thalf\right)^K \,p\, \indct{\bigvee_{i=1}^K (a_{t,i} = j)} \sgn(x_{t,j}y_t) }\right)
\\
&=  \sum_{\r} \left(\thalf\right)^{(K+1)T} \sum_{t=1}^T
\indct{\bigvee_{i=1}^K (a_{t,i} = j)} \Bigl(- \log\bigr(1 + 2 \,p ~\sgn(x_{t,j}y_t)\bigr)\Bigr)
\\
&=  \sum_{t=1}^T \Eu\left[
\indct{\bigvee_{i=1}^K (a_{t,i} = j)} \Bigl(- \log\bigl(1 + 2 \,p ~\sgn(x_{t,j}y_t)\bigr)\Bigr) \right]
\\
&=  \sum_{t=1}^T P_u\left[\bigvee_{i=1}^K (a_{t,i} = j)\right]
\Eu\Bigl[- \log\bigl(1 + 2 \,p ~\sgn(x_{t,j}y_t)\bigr) \Bigr]
\\ &\quad
    \text{(since $x_{t,j}y_t$ is independent of $a_{t,1},\dots,a_{t,K}$)}
\\
&=  \Bigl(\thalf(- \log\bigl(1 + 2 \,p)\bigr) + \thalf\bigl(- \log(1 - 2 \,p)\bigr) \Bigr) 
\sum_{t=1}^T P_u\left[\bigvee_{i=1}^K (a_{t,i} = j)\right]
\\
&= - \thalf \log(1-4p^2) \Eu[N_j] ~.
\end{split}
\]
Combining all the above we conclude our proof. 
\end{proof}

We have shown previously that the expected reward of any algorithm is
bounded above by $\frac{1}{2}T + p \Ej[N_j]$. Applying \lemref{lem:A1} above on $f(\r) = N_j \in \{0,1,\dots,T\}$ we get that 
\[
    \Ej[N_j] \le \Eu[N_j] + T \sqrt{- \log(1-4p^2)\Eu[N_j]  } ~.
\]
Therefore, the expected reward of any algorithm is at most
\[
    \tfrac{1}{2}T + p \left(\Eu[N_j] +  T \sqrt{- \log(1-4p^2)\Eu[N_j]  } \right)~.
\]
Since the adversary will choose $j$ to minimize the above and since
the minimum over $j$ is smaller then the expectation over choosing $j$
uniformly at random we have that the reward against an adversarial
choice of $j$ is at most
\begin{equation} \label{eqn:banpr1}
    \tfrac{1}{2}T + p \, \frac{1}{d} \sum_{j=1}^d \left(\Eu[N_j] +  T \sqrt{- \log(1-4p^2)\Eu[N_j]  } \right)~.
\end{equation}
Note that 
\[
    \frac{1}{d} \sum_{j=1}^d \Eu[N_j] = \frac{1}{d} \Eu[N_1+\ldots+N_d] \le \frac{KT}{d} ~.
\]
Combining this with \eqref{eqn:banpr1} and using Jensen's inequality
we obtain the following upper bound on the reward
\[
    \tfrac{1}{2}T + p \left(\tfrac{K}{d}T + T \sqrt{-\log(1-4p^2) \tfrac{K}{d}T} \right) ~.
\]
Assuming that $\epsilon \le 1/16$ we have that $4p^2 = 4\epsilon \le
1/4$ and thus using the inequality $-\log(1-q) \le \tfrac{3}{2}q$, which
holds for $q \in [0,1/4]$, we get the upper bound
\begin{equation} \label{eqn:upperStrategy}
    \tfrac{1}{2}T + p \left(\tfrac{K}{d}T + T\sqrt{\tfrac{6K}{d}p^2T} \right) ~.
\end{equation}


\paragraph{Concluding the proof:}
Take a learning algorithm that finds an $\epsilon$-good predictor
using $m$ examples. Since the reward of the strategy based on this
learning algorithm cannot exceed the upper bound given in
\eqref{eqn:upperStrategy} we obtain that:
\[
    \tfrac{1}{2}T + (T-m)p \le  \tfrac{1}{2}T + p \left(\tfrac{K}{d}T + T\sqrt{\tfrac{6K}{d}p^2T} \right)\]
which solved for $m$ gives
\[
    m \ge T \left(1 - \tfrac{K}{d} - \sqrt{\tfrac{6K}{d}p^2T} \right)~.
\]
Since we assume $d \ge 4K$, choosing $T = \left\lfloor d \big/ (96 Kp^2)\right\rfloor$,
and recalling $p^2 = \epsilon$, gives
\[
    m \ge \frac{T}{2} = \frac{1}{2}\left\lfloor\frac{d}{96 K\epsilon}\right\rfloor ~.
\]


\subsection{Proof of \thmref{thm:lowertest}}

Let $\w^\star = (1/3,1/3,1/3)$. Let $\x \in \{\pm 1\}^3$ be distributed
uniformly at random and $y$ is determined deterministically to be
$\inner{\w^\star,\x}$.  Then, $L_\D(\w^\star) = 0$. However, 
any algorithm that only view $2$ attributes have an
uncertainty about the label of at least $\pm \tfrac{1}{3}$, and
therefore its expected squared error is at least $1/9$. Formally,
suppose the algorithm asks for the first two attributes and form its
prediction to be $\hat{y}$. 
Since the generation of attributes is independent, we have that the
value of $x_3$ does not depend on $x_1,x_2$, and $\hat{y}$, and 
therefore 
\[
\E[ (\hat{y} - \inner{\w^\star,\x})^2 ] = 
\E[ (\hat{y} - w^\star_1x_1 - w^\star x_2 - w^\star_3 x_3)^2] =
\E[ (\hat{y} - w^\star_1x_1 - w^\star x_2 )^2]  + 
\E[ ( w^\star_3 x_3)^2] \ge 0 + (1/3)^2 \E[x_3^2] = 1/9 ~,
\]
which concludes our proof.

\end{document}